\documentclass[preprint,11pt]{elsarticle}
\usepackage{graphicx}
\usepackage{setspace}
\usepackage{amsmath}
\usepackage{bbm}
\usepackage{dsfont}
\usepackage{amsfonts}
\usepackage{lipsum}
\usepackage{multicol}
\usepackage{float}
\usepackage{bm}
\usepackage{color}
\usepackage{etoolbox}
\usepackage{soul}
\usepackage[linesnumbered,ruled,vlined]{algorithm2e}
\usepackage{mathtools}
\usepackage{caption}
\usepackage{amsthm}
\usepackage{adjustbox}
\usepackage{afterpage}
\usepackage{mathrsfs}
\usepackage{oplotsymbl}
\usepackage{booktabs}
\usepackage{multirow}
\usepackage{array}
\usepackage{xcolor}
\usepackage{amssymb}
\usepackage{threeparttable}
\usepackage[caption = false,subrefformat=parens,labelformat=parens]{subfig}
\usepackage{hyperref}

\begin{document}

\begin{frontmatter}



\title{Forecasting-based Biomedical Time-series Data Synthesis for Open Data and Robust AI}

\author[kaist-ee]{Youngjoon Lee}
\author[kaist-ee]{Seongmin Cho}
\author[kaist-ee]{Yehhyun Jo}
\author[hansung]{Jinu Gong}
\author[kaist-ee,kaist-bbe]{\\[1ex]Hyunjoo Jenny Lee}
\author[kaist-ee]{Joonhyuk Kang\corref{cor1}}

\cortext[cor1]{Corresponding Author: \texttt{jkang@kaist.ac.kr}}

\address[kaist-ee]{School of Electrical Engineering, KAIST, Daejeon, Republic of Korea}
\address[hansung]{Department of Applied AI, Hansung University, Seoul, Republic of Korea}
\address[kaist-bbe]{Department of Bio and Brain Engineering, KAIST, Daejeon, Republic of Korea}

%% Abstract
\begin{abstract}
The limited data availability due to strict privacy regulations and significant resource demands severely constrains biomedical time-series AI development, which creates a critical gap between data requirements and accessibility. 
Synthetic data generation presents a promising solution by producing artificial datasets that maintain the statistical properties of real biomedical time-series data without compromising patient confidentiality.
While GANs, VAEs, and diffusion models capture global data distributions, forecasting models offer inductive biases tailored for sequential dynamics.
We propose a framework for synthetic biomedical time-series data generation based on recent forecasting models that accurately replicates complex electrophysiological signals such as EEG and EMG with high fidelity. 
These synthetic datasets can be freely shared for open AI development and consistently improve downstream model performance.
Numerical results on sleep-stage classification show up to a 3.71\% performance gain with augmentation and a 91.00\% synthetic-only accuracy that surpasses the real-data-only baseline.
\end{abstract}

% \begin{figure}[h]
%     \centering
%     \includegraphics[width=\textwidth]{figure/graphical_abstract.pdf}
% \end{figure}

% \begin{highlights}
% \item The synthetic data generation is achieved through a novel forecasting-based method that trains on real biomedical signals and produces high-fidelity datasets.
% \item Open-source contribution is demonstrated as synthetic datasets are shared on public repositories, ensuring ease of use while maintaining privacy compliance.
% \item Data gap fulfillment is accomplished as synthetic samples supplement underrepresented regions in real datasets and provide a more complete data distribution for AI training in closed environments.
% \end{highlights}

%% Keywords

\end{frontmatter}



\newpage

\section{Introduction}\label{intro}

\begin{figure}[t]
    \centering
    \includegraphics[width=\textwidth]{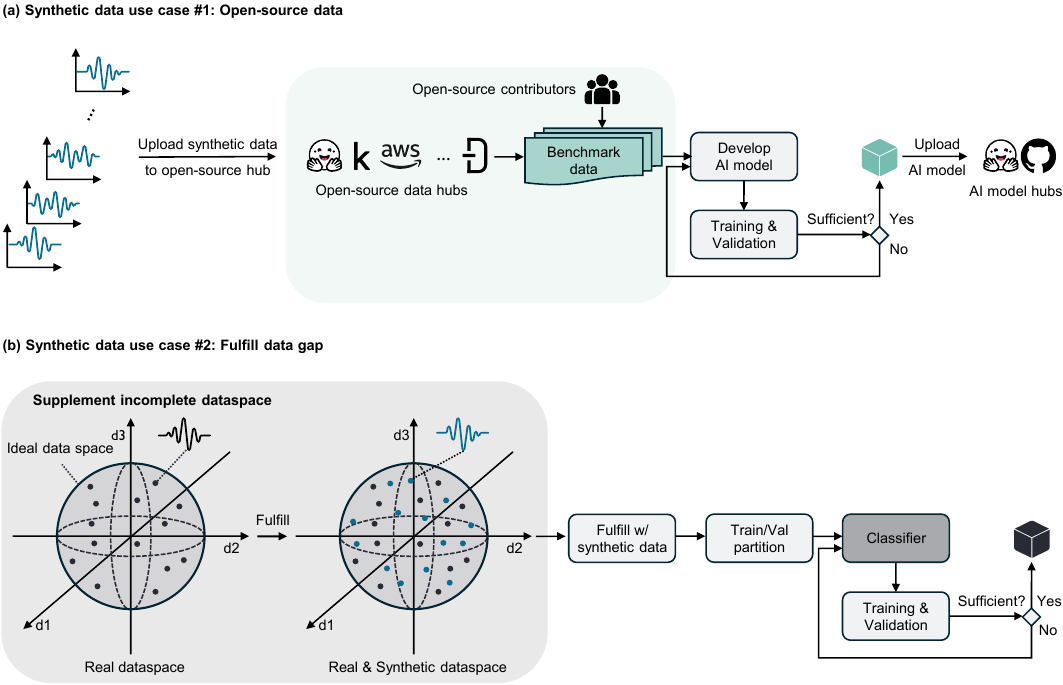}
    \caption{\textbf{Open-source contribution and data gap fulfillment.} (a) Open-source data contribution: Synthetic datasets are uploaded to public repositories (Hugging Face, Kaggle, AWS), enabling broader access for open AI development while maintaining privacy compliance. (b) Data gap fulfillment: Synthetic samples populate underrepresented regions in the feature space of real datasets, enhancing classifier training by providing a more comprehensive representation of the ideal data distribution.
    }
    \label{fig:0}
\end{figure}

The development of high-performance AI models for biomedical time-series applications requires extensive and diverse datasets \cite{acosta2022multimodal,zhou2023foundation}.
Biomedical data availability, however, remains severely constrained due to stringent privacy regulations, substantial acquisition costs, and the rarity of certain biomedical conditions \cite{moor2023foundation}. 
These constraints create a considerable disparity between the data necessary for robust AI development and the data currently accessible to researchers \cite{mahmood2025benchmarking}. 
Synthetic data generation offers a powerful strategy for addressing these limitations through the creation of artificial datasets that maintain the statistical properties of real biomedical data without compromising patient confidentiality \cite{chen2021synthetic}. 
This approach significantly advances biomedical AI research by directly mitigating critical data availability challenges that currently impede progress in AI-driven applications \cite{van2024synthetic}.

Biomedical signals contain complex inter-signal patterns that must be retained in synthetic data to ensure biomedical relevance and practical utility \cite{norori2021addressing}.
Moreover, the variability of temporal dynamics across signal types poses additional challenges for synthetic data generation.
The time-series forecaster methods excel at understanding the temporal characteristics of sequential data better than any other models \cite{torres2021deep}. 
These techniques identify and replicate underlying data patterns to generate realistic continuations of time signals, preserving essential temporal properties vital for diagnostic and monitoring applications \cite{chen2023long}. 
Notably, this methodology demonstrates exceptional adaptability across diverse biomedical signal types, including electromyography (EMG), electroencephalograms (EEGs), and continuous glucose monitoring data, each with unique temporal signatures that forecasting methods can effectively capture and reproduce. 
Thus, the resulting synthetic data derived from time-series forecasting maintains high biomedical relevance and accuracy required for meaningful applications.

The integration of synthetic biomedical data into open-source repositories substantially accelerates research and innovation in biomedical-focused AI \cite{mahmood2025benchmarking}. 
Many current open-source biomedical datasets lack sufficient size and diversity, severely restricting the development of robust and generalizable AI models. 
Contributing high-quality synthetic datasets to prominent platforms such as Hugging Face, Kaggle, AWS, and data.gov significantly expands available resources while democratizing data access across the biomedical AI research community. 
Moreover, the synthetic datasets inherently protect patient privacy by generating artificial samples without personally identifiable information, while adherence to international privacy standards, including the General Data Protection Regulation (GDPR) \cite{regulation2018general}, enables unrestricted global data sharing and eliminates risks of patient re-identification \cite{yoon2020anonymization}.

In addition, absolute data scarcity represents a critical challenge for biomedical AI model development, particularly for rare diseases or emerging biomedical conditions where existing real-world datasets frequently lack sufficient data points to train robust and reliable AI models \cite{bansal2022systematic}.
We address these critical shortages through strategic deployment of synthetic data generation to produce additional relevant samples that mirror the characteristics of limited real datasets \cite{chen2024towards,loecher2021using}. 
The AI models trained on these expanded datasets consistently demonstrate improved predictive performance and enhanced generalization capabilities \cite{PEZOULAS20242892}. 
This synthetic data supplementation directly alleviates the issue of insufficient biomedical data availability, substantially advancing the effectiveness and applicability of AI systems in biomedical environments \cite{gao2023synthetic}.

Despite the strengths of GANs \cite{yoon2019time,esteban2017real}, VAEs \cite{desai2021timevae,pmlr-v206-lee23d}, and diffusion models \cite{tashiro2021csdi,timeweaver2024} in general data synthesis, the sequential nature of biomedical time signals requires models that explicitly incorporate temporal inductive biases. 
Here, we introduce a paradigm-shifting approach that repurposes time-series forecasters—used for next-step prediction—as synthesizers to address the persistent challenges of biomedical AI: privacy and data scarcity (Fig. \ref{fig:0}).
By leveraging forecasting models, our approach generates high-fidelity sequences that better capture real-world signal dynamics than general-purpose baselines.
To validate our approach, we synthesized EEG sleep-stage signals using 16 state-of-the-art time-series forecasting models \cite{dialtedrnn,tcn,nbeats,deepar,tft,bitcn,nbeatsx,nhits,dlinear,patchtst,timesnet,tide,deepnpts,itransformer,softs,kan}. 
When combined, synthetic data consistently improved performance across all subjects, with DLinear achieving a 3.71 percentage point gain. 
Moreover, synthetic-only training slightly outperformed original data, with SOFTS achieving 91.00\% compared to 90.83\%. 
The results indicate that forecasting-based synthesis is a practical approach that supports privacy-preserving data sharing and improves model robustness by enriching underrepresented regions.

The remainder of this paper is organized as follows. 
Section~\ref{method} details the animal data acquisition procedure, the forecasting-based synthesis pipeline, and the evaluation framework.
Section~\ref{result} reports the experimental findings, including data preparation, performance and class-wise analyses, visualization of original versus synthetic data, quantitative quality and privacy assessments, and comparison with representative GAN-based synthesis.
Section~\ref{discussion} analyzes the effect of the synthetic-to-original data ratio on performance, highlights key limitations, and proposes directions for future work.
Finally, Section~\ref{conclusion} provides concluding remarks.

\section{Material and Methods}\label{method}
\subsection{Animal Data Acquisition}
All \textit{in vivo} data was acquired from a previous study (IACUC\footnote{Institutional Animal Care and Use Committee} approval: KA2021-066) \cite{jo2022general}. 
Briefly, EEG/EMG signals were recorded at a sampling rate of 1 kHz using a biopotential acquisition device (RHD2000, Intan Technologies, CA, USA), which was then amplified and digitally filtered (low-pass filter at 0.1 Hz, high-pass filter at 7.5 kHz, and notch filter at 60 Hz). 
Filtered EEG/EMG signals were segmented into 5s epochs for further analysis using custom-written software (MATLAB, MathWorks, Natick, MA, USA).

\begin{figure}[t]
    \centering
    \includegraphics[width=\textwidth]{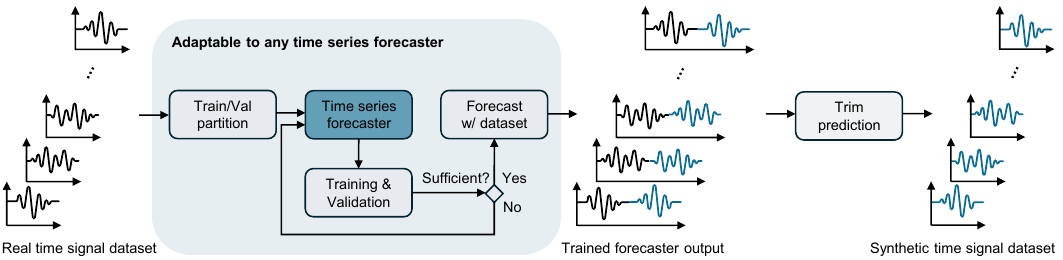}
    \caption{\textbf{Framework for synthetic biomedical time-series data generation.} Synthetic data generation using time-series forecasters: The forecasting model trained on real biomedical signals generates similar patterns or enriches underrepresented segments.
    }
    \label{fig:1}
\end{figure}

\subsection{Synthetic Data Generation}
The synthetic signals were generated from real EEG recordings using a class-conditional time-series forecasting approach (Fig.~\ref{fig:1}). 
Formally, let $D=\{(y^{(i)}, c^{(i)})\}_{i=1}^N$ denote the original dataset, where $y^{(i)}$ represents an EEG time-series, and $c^{(i)} \in \mathcal{C}$ is the corresponding sleep stage label from the class set $\mathcal{C}=\{\text{``WAKE''}, \text{``NREM''}, \text{``REM''}\}$.

For each class $c \in \mathcal{C}$, a separate time-series forecasting model $f_c(\cdot;\theta_c)$ was trained using supervised input--output pairs generated with a sliding window of size $L$.
The forecaster is designed to predict a future segment of length $L$, acting as a building block to construct a complete sequence of total length $H$ (e.g., $H=500$).
Training pairs were constructed as:
\begin{equation}
(x_t, y_t) = \big(y^{(i)}_{t:t+L-1}, \; y^{(i)}_{t+L:t+2L-1}\big),
\end{equation}
where $x_t \in \mathbb{R}^L$ is the context window and $y_t \in \mathbb{R}^L$ is the target segment to be predicted.
For all experiments, we varied $L \in \{10,25,50,100,250\}$.
Model parameters were optimized using the Huber loss \cite{meyer2021alternative}:
\begin{equation}
\mathcal{L}_c(\theta_c)
= \frac{1}{|D_c|} \sum_{(x_t,y_t) \in D_c} 
\mathcal{L}_{\text{Huber}} \!\big(f_c(x_t;\theta_c), y_t\big),
\end{equation}
where $D_c = \{(y^{(i)},c^{(i)}) \in D \mid c^{(i)} = c\}$.
All models were trained with batch size 32 and a maximum of 1,000 optimization steps at a sampling rate of 100 Hz (10 ms interval).

After training, each model $f_c$ was used to generate synthetic EEG signals through a sliding window reconstruction strategy.
Instead of relying solely on previously generated predictions, the model used temporally aligned context extracted from the original signal to maintain stable temporal structure, while generating new model-based variations at each step.
Given a context window $x^{(i)} \in \mathbb{R}^L$ sampled from $y^{(i)}$, the model iteratively predicts non-overlapping $L$-step segments:
\begin{equation}
\tilde{y}^{(i)}_k = f_{c^{(i)}}(x^{(i)}_k;\theta_{c^{(i)}}^\ast),
\end{equation}
which are sequentially concatenated until the desired length $H$ is reached.
The final synthetic dataset is defined as:
\begin{equation}
\hat{D} = \{(\hat{y}^{(i)}, c^{(i)})\}_{i=1}^N, \qquad \hat{y}^{(i)} \in \mathbb{R}^H.
\end{equation}
Algorithm~\ref{alg:1} summarizes the synthesis procedure.

\begin{algorithm}[t]
\caption{Forecasting-based Time-series Data Synthesis}
\label{alg:1}
\DontPrintSemicolon

\BlankLine
\textbf{Train Forecasters}\;
\ForEach{$c \in \mathcal{C}$}{
    Construct $D_c=\{(x_t,y_t)\}$ via sliding windows\;
    $\theta_c^\ast \leftarrow 
    \arg\min_{\theta_c}
    \frac{1}{|D_c|}\!
    \sum_{(x_t,y_t)\in D_c}
    \mathcal{L}_{\text{Huber}}\big(f_c(x_t;\theta_c), y_t\big)$\;
}

\BlankLine
\textbf{Generate Synthetic Signals}\;
\ForEach{$(y^{(i)},c^{(i)})$}{
    $\hat{y} \leftarrow [\,]$\;
    \For{$k=1$ \KwTo $H/L$}{
        Extract context $x^{(i)}_k$ of length $L$ from $y^{(i)}$\;
        $\tilde{y}^{(i)}_k \leftarrow f_{c^{(i)}}(x^{(i)}_k;\theta_{c^{(i)}}^\ast)$\;
        Append $\tilde{y}^{(i)}_k$ to $\hat{y}$\;
    }
    Add $(\hat{y},c^{(i)})$ to $\hat{D}$\;
}
\Return $\hat{D}$\;
\end{algorithm}

\subsection{Evaluation Framework}
To evaluate synthetic data utility, we designed a classification framework with three training conditions: original data only (O), synthetic data only (S), and combined original and synthetic data (O+S).
Let $D_{\text{orig}}=\{(y^{(i)},c^{(i)})\}_{i=1}^N$ and
$D_{\text{syn}}=\{(\hat{y}^{(i)},c^{(i)})\}_{i=1}^N$ represent the original and synthetic datasets.
In the O+S condition, the training set was constructed as:
\begin{equation}
D_{\text{train}} = D_{\text{orig}} \cup D_{\text{syn}}.
\end{equation}
Each time-series sample $u^{(i)} \in \mathbb{R}^H$ (representing either a real $y^{(i)}$ or synthetic $\hat{y}^{(i)}$ signal) was converted into a time-frequency representation using the short-time Fourier transform (STFT):
\begin{equation}
S^{(i)}(f,t) = \left| \sum_{\tau=0}^{H-1} u^{(i)}(\tau)\,w(\tau-t) \,e^{-j2\pi f\tau} \right|^2,
\end{equation}
where $w(\cdot)$ represents a Hann window of 128 points with 50\% overlap.
The resulting spectrograms were transformed using a logarithmic scale $\log(1+S^{(i)})$ and standardized to zero mean and unit variance.
These spectrograms served as input, adapted for time-series classification by setting the input channel to one and the output to the number of sleep stage classes.

Model training was performed on a fixed train/test split, with training and testing sets predefined for each subject.
The classifier model was a ResNet-18 \cite{he2016deep}, trained using stochastic gradient descent with a learning rate of $10^{-4}$.
Note that while ResNet-18 served as the default classifier, the framework is compatible with other image classification models.
To ensure reproducibility, all runs were executed with five random seeds, using AMD Instinct MI300X GPUs supported by AMD Developer Cloud credits.

Let $f_\theta(\cdot)$ denote the classifier parameterized by weights $\theta$.
The objective function was the cross-entropy loss:
\begin{equation}
\mathcal{L}(\theta) = - \sum_{i} \sum_{k=1}^{|\mathcal{C}|} \mathbbm{1}[c^{(i)} = k] \log p_\theta^{(i)}(k),
\end{equation}
where $\mathbbm{1}[\cdot]$ is the indicator function, $c^{(i)}$ is the ground-truth class label, $p_\theta^{(i)}(k)$ represents the predicted probability for class $k$ on sample $i$, and $|\mathcal{C}|$ denotes the number of classes.
The original-only (O) and synthetic-only (S) conditions followed the same preprocessing and evaluation procedures as the O+S condition, ensuring consistency across all experimental settings.

\subsection{Performance Metrics}
Model performance was quantitatively assessed using accuracy and F1-score across all training conditions. Let $c^{(i)} \in \mathcal{C}$ and $\hat{c}^{(i)} \in \mathcal{C}$ denote the ground-truth and predicted class labels for the $i$-th sample in the test set.
The primary evaluation metric was classification accuracy, computed as:
\begin{equation}
\text{Accuracy} = \frac{1}{N_{\text{test}}} \sum_{i=1}^{N_{\text{test}}} \mathbbm{1}[c^{(i)} = \hat{c}^{(i)}],
\end{equation}
where $N_{\text{test}}$ represents the total number of test samples.

To evaluate class-wise performance, we computed F1-score for each class $k \in \mathcal{C}$. Let $\text{TP}_k$, $\text{FP}_k$, and $\text{FN}_k$ denote the number of true positives, false positives, and false negatives for class $k$. Precision and recall were defined as:
\begin{equation}
\text{Precision}_k = \frac{\text{TP}_k}{\text{TP}_k + \text{FP}_k}, \quad \text{Recall}_k = \frac{\text{TP}_k}{\text{TP}_k + \text{FN}_k}.
\end{equation}
The F1-score, representing the harmonic mean of precision and recall, was computed as:
\begin{equation}
\text{F1}_k = 2 \cdot \frac{\text{Precision}_k \cdot \text{Recall}_k}{\text{Precision}_k + \text{Recall}_k}.
\end{equation}

\section{Results}\label{result}
\subsection{Data Preparation}
\begin{figure}[t]
    \centering
    \includegraphics[width=\textwidth]{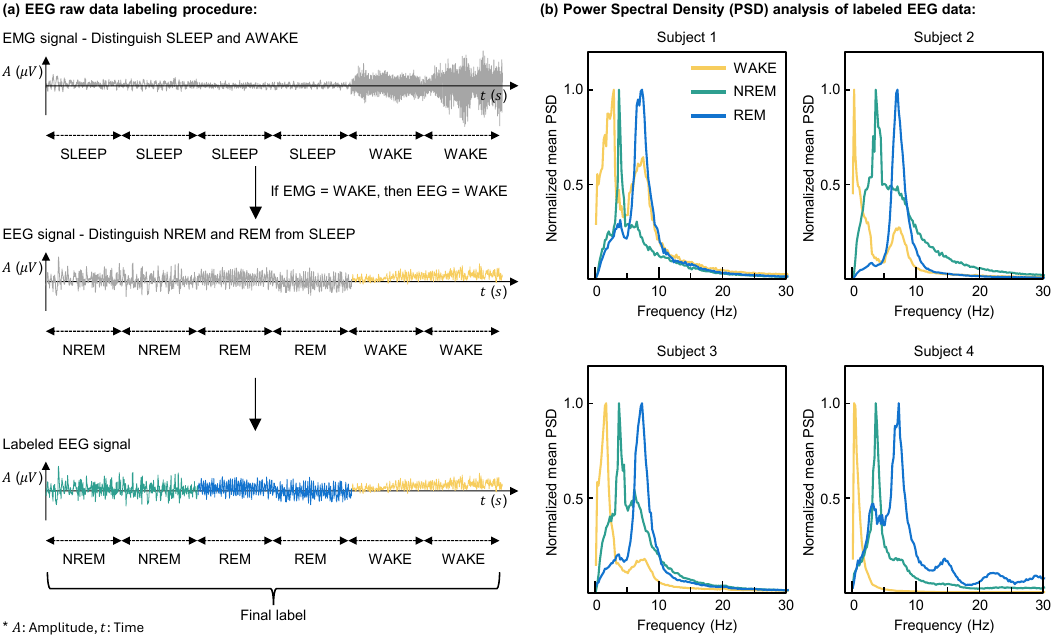}
    \caption{\textbf{The two-stage EEG data labeling procedure and corresponding power spectral density (PSD) analysis.} (a) EEG raw data labeling process using EMG signals to differentiate between WAKE and SLEEP states, followed by EEG frequency domain filtering to distinguish NREM and REM sleep stages. (b) Normalized mean PSD analysis demonstrates distinct spectral signatures across four different subjects, validating the effectiveness and consistency of the labeling process.
    }
    \label{fig:2}
\end{figure}
The EEG dataset underwent a two-stage labeling process to achieve accurate sleep stage identification following methods described in \cite{jo2022general} (Fig. \ref{fig:2}).
Initially, EMG signals served as reference signals to differentiate between SLEEP and WAKE states by comparing EMG power against a baseline threshold. 
Subsequently, epochs identified as SLEEP underwent EEG signal analysis using frequency domain filtering via fast Fourier transform (FFT) to characterize sleep stages with high specificity. 
We intentionally filtered frequency bands to highlight pronounced peaks within the delta frequency range (0.5-4 Hz) for NREM sleep and distinct peaks within the theta frequency range (4-8 Hz) for REM sleep.

In particular, NREM sleep is characterized by an increase in delta wave activity, while strong theta wave power is observed during REM sleep. 
Then, we filtered the frequency bands to highlight pronounced peaks within the delta frequency range (0.5-4 Hz) for NREM sleep and distinct peaks within the theta frequency range (4-8 Hz) for REM sleep.
This labeled EEG dataset provide the foundation for synthetic data generation using time-series forecasting models.

\afterpage{
\newpage

\begin{table*}[t]
\centering
\caption{\textbf{The two-stage EEG data labeling Test accuracy (\%) comparison for different time-series generators across all subjects.} O: Only original data S: Only synthetic data, O+S: Original with synthetic data.}
\label{tab:1}
\resizebox{\textwidth}{!}{
\begin{tabular}{lcccccccc}
\toprule
\multirow{2}{*}{Time-series Forecaster} &
\multicolumn{2}{c}{Subject 1 (O: 90.83\%)} &
\multicolumn{2}{c}{Subject 2 (O: 96.77\%)} &
\multicolumn{2}{c}{Subject 3 (O: 95.50\%)} &
\multicolumn{2}{c}{Subject 4 (O: 86.60\%)} \\
\cmidrule(lr){2-3}\cmidrule(lr){4-5}\cmidrule(lr){6-7}\cmidrule(lr){8-9}
& S & O+S & S & O+S & S & O+S & S & O+S \\
\midrule
Dilated RNN~\cite{dialtedrnn} & 84.08 & 91.08 & 90.53 & 97.12 & 87.81 & 96.53 & 71.32 & 89.06 \\
TCN~\cite{tcn}               & 85.23 & 92.16 & 92.94 & 97.44 & 87.89 & 96.33 & 71.07 & 89.37 \\
N-BEATS~\cite{nbeats}         & 89.58 & 92.33 & 96.57 & 97.55 & 94.71 & 96.73 & 68.62 & 90.00 \\
DeepAR~\cite{deepar}          & 78.08 & 92.75 & 94.87 & 97.43 & 93.61 & 96.61 & 56.67 & 88.74 \\
TFT~\cite{tft}                & 82.50 & 91.91 & 90.26 & 97.32 & 84.70 & 96.49 & 71.82 & 89.87 \\
BiTCN~\cite{bitcn}            & 90.00 & 92.50 & 96.45 & 97.67 & 94.48 & 96.53 & 59.25 & 89.68 \\
NBEATSx~\cite{nbeatsx}        & 89.58 & 92.33 & 96.57 & 97.55 & 94.71 & 96.73 & 68.62 & 90.00 \\
N-HiTS~\cite{nhits}           & 89.25 & 92.50 & 96.53 & 97.51 & 94.20 & 96.52 & 71.01 & 90.13 \\
DLinear~\cite{dlinear}        & 86.58 & 91.67 & 94.60 & 97.08 & 84.18 & 95.94 & 63.14 & 90.31 \\
PatchTST~\cite{patchtst}      & 85.25 & 91.75 & 88.91 & 97.12 & 89.11 & 96.61 & 70.38 & 88.62 \\
TimesNet~\cite{timesnet}      & 86.08 & 91.00 & 92.23 & 97.28 & 88.60 & 96.49 & 72.14 & 89.18 \\
TiDE~\cite{tide}              & 89.83 & 92.42 & 96.76 & 97.71 & 94.60 & 96.96 & 61.76 & 89.12 \\
DeepNPTS~\cite{deepnpts}      & 88.92 & 91.33 & 92.43 & 97.36 & 94.40 & 96.61 & 70.44 & 88.99 \\
iTransformer~\cite{itransformer} & 89.92 & 91.08 & 95.50 & 97.04 & 94.32 & 96.33 & 69.56 & 87.86 \\
SOFTS~\cite{softs}            & 91.00 & 92.00 & 94.68 & 97.24 & 95.23$^\ast$ & 96.65 & 82.08 & 89.62 \\
KAN~\cite{kan}                & 90.16 & 92.09 & 96.49 & 97.63 & 95.23$^\ast$ & 96.69 & 70.57 & 88.68 \\
\bottomrule
\end{tabular}
}
\begin{tablenotes}
\scriptsize
\item $^\ast$ Ties at 95.23 for Subject 3 in SOFTS and KAN.
\end{tablenotes}
\end{table*}

\begin{figure}
    \centering
    \includegraphics[width=\textwidth]{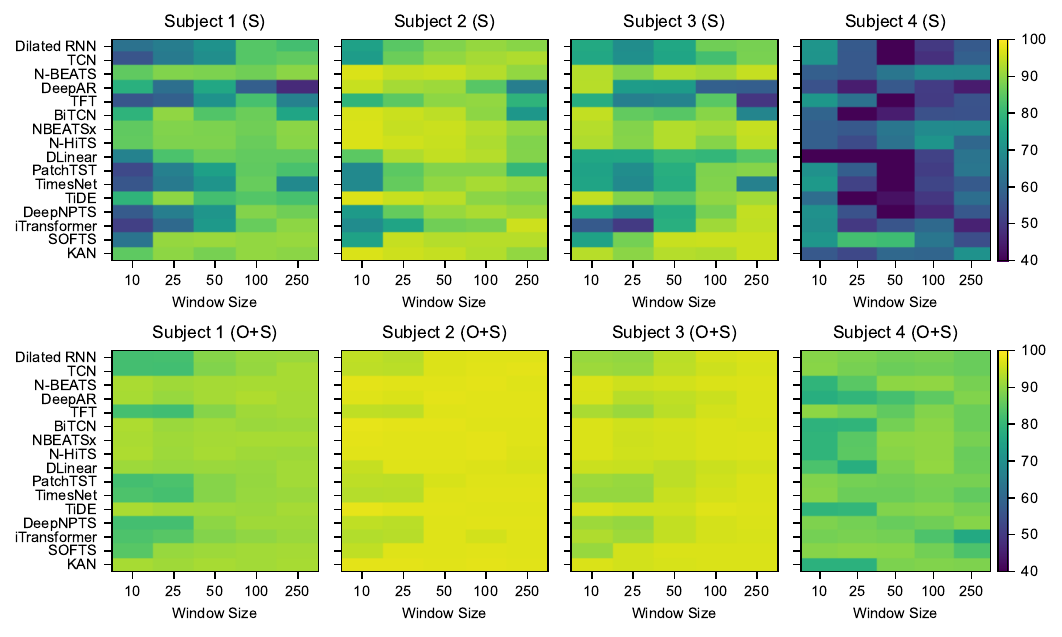}
    \caption{\textbf{Performance heatmap across window sizes (10, 25, 50, 100, 250) for 16 time-series forecasters evaluated on each subject.} (a) The synthetic-only (S) condition is shown in the top row, and (b) the combined original with synthetic data (O+S) condition in the bottom row. Lighter colors indicate higher accuracy, reflecting better model performance.
    }
    \label{fig:3}
\end{figure}
\clearpage
}

\subsection{Performance Analysis}
We evaluated 16 state-of-the-art time-series forecasting models for synthetic biomedical data generation across four subjects (Table \ref{tab:1}). 
The models in the Transformer family (SOFTS, TiDE, iTransformer, TFT, PatchTST, TimesNet) consistently achieved the highest accuracy across all subjects, demonstrating strong capacity to model long-range dependencies inherent in biomedical signals.
As \cite{trirat2024universal}, this suggests that the global receptive field of self-attention mechanisms is well suited for modeling long-range dependencies.
Notably, MLP-based models (N-BEATS, NBEATSx, N-HiTS, DLinear, DeepNPTS) also exhibited robust performance, particularly at larger window sizes, highlighting their ability to capture essential signal structures. 
Meanwhile, RNN-based models (DilatedRNN, DeepAR) contributed positively, though their performance varied with subject characteristics. 
The CNN-based models (TCN, BiTCN) showed competitive accuracy in several settings. 
Additionally, KAN, belonging to the ‘any’ category, ranked among the top performers in multiple cases. 
The synthetic-only (S) training condition frequently approached original-only accuracy, indicating the high fidelity of the generated data. 
Thus, the open release of these synthetic datasets through platforms such as Hugging Face and Kaggle is expected to further accelerate progress in biomedical AI.

When combining synthetic data with original data (O+S), accuracy consistently improved across all models and subjects. 
The TiDE achieved the highest accuracy overall, reaching 97.71\% for Subject 2, while DeepAR obtained 92.75\% for Subject 1. 
For Subject 3, TiDE reached 96.96\%, surpassing the original-only baseline. The largest improvement was observed for Subject 4, where DLinear increased from 63.14\% (S) to 90.31\% (O+S), demonstrating the complementary value of synthetic augmentation. 
The Transformer-based models maintained leading performance across subjects, while models from other families also achieved notable gains. 
Moreover, incorporating synthetic data reduced variability across models, suggesting improved stability and generalization. 
These findings confirm that forecasting-based synthetic data effectively enhances performance in biomedical time-series applications.

In addition, the test accuracy patterns across varying window sizes and model families reveal further insights (Fig. \ref{fig:3}). 
In the S condition, larger window sizes generally yielded higher accuracy, with Transformer and MLP models maintaining strong and consistent performance across settings. 
By contrast, the CNN and RNN models exhibited greater sensitivity to window size, reflecting their dependence on local temporal patterns. 
Under the O+S condition, the test accuracy improved substantially across all subjects, and variability across window sizes was significantly reduced. 

\begin{figure}
    \centering
    \includegraphics[width=\textwidth]{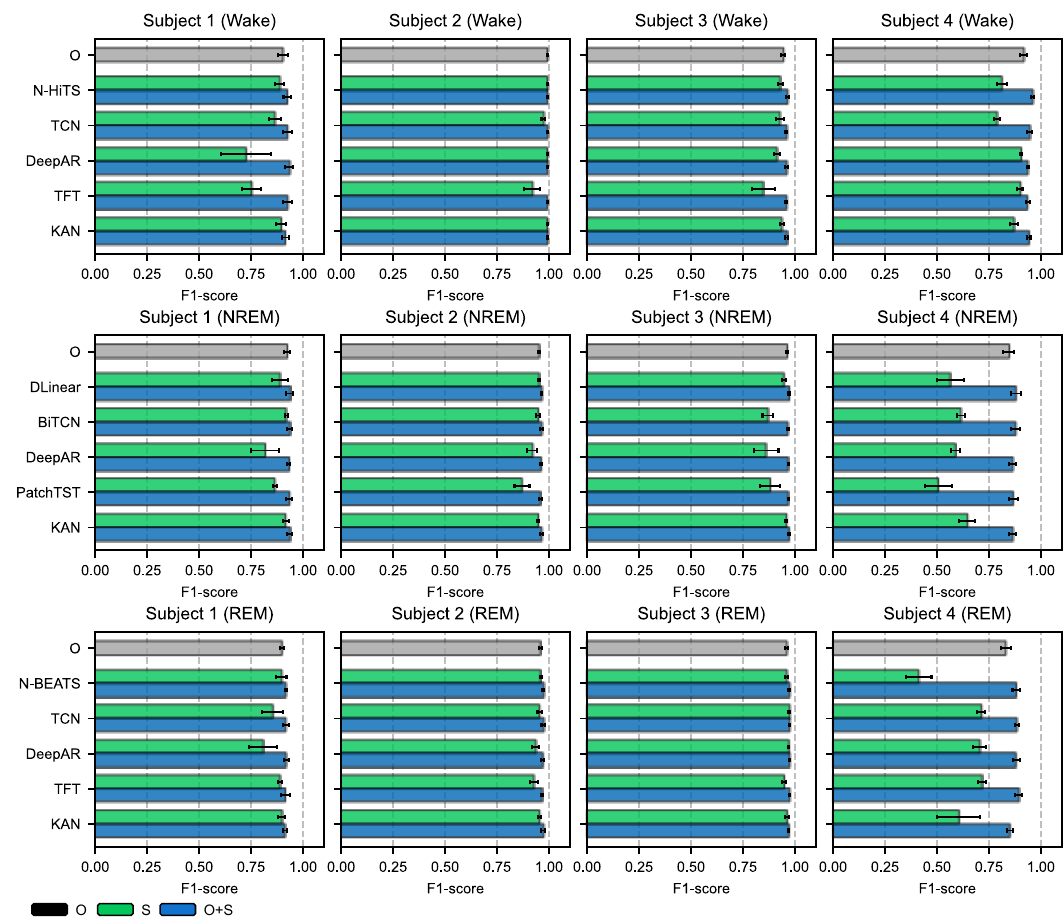}
    \caption{\textbf{Class-wise F1-scores across subjects for the synthetic-only (S) and combined original with synthetic data (O+S) conditions.} For each model family, the best-performing forecaster was selected separately for S and O+S settings. All values represent mean ± std computed across 5 random seeds.
    }
    \label{fig:4}
\end{figure}

\subsection{Class-wise Performance Analysis}
For each subject, class-wise performance was evaluated under both the original-only (O) and combined (O+S) conditions, using the best-performing forecaster from each model family for both S and O+S settings (Fig. \ref{fig:4}). 
Under the synthetic-only (S) condition, Transformer and MLP models generally produced higher class-wise F1-scores compared to RNN and CNN families. 
For example, in Subject 1, KAN (any family) achieved 0.893 (Wake) and 0.896 (REM), while BiTCN reached 0.919 for NREM. 
In Subject 2, TiDE (Transformer family) yielded nearly perfect F1-scores across all classes, including 0.992 (Wake) and 0.960 (REM). 
Similarly, Subject 3 showed strong S-only performance with KAN and BiTCN models achieving F1-scores of 0.936 (Wake), 0.957 (NREM), and 0.973 (REM). Subject 4, though more challenging, demonstrate solid performance in Wake (0.905, DilatedRNN) and REM (0.715, TFT), with greater variability in NREM across families. 

When synthetic data was combined with original data (O+S), F1-scores improved consistently across all families, subjects, and classes. 
In Subject 1, DeepAR and DLinear models contributed to strong gains, achieving 0.933 (Wake), 0.936 (NREM), and 0.919 (REM). 
The Subject 2 demonstrated further gains, with TiDE and TCN models producing near-ceiling F1-scores—0.994 (Wake), 0.965 (NREM), and 0.973 (REM). 
In Subject 3, O+S training pushed F1-scores above original baselines for REM (0.973 with TimesNet and BiTCN), while maintaining high values for Wake (0.965) and NREM (0.971). 
In Subject 4, where S-only variability was greater, O+S training brought substantial improvements across all classes, achieving 0.960 (Wake), 0.881 (NREM), and 0.894 (REM). 
Notably, gains were observed consistently across Transformer, MLP, RNN, and CNN families, with Transformer-based models contributing prominently in harder classes such as REM. 

These numerical results show that Transformer and MLP forecasters are particularly effective in synthesizing biomedical signals with fine-grained class fidelity, while RNN and CNN families contribute targeted advantages in specific conditions. 
When combined with original data, synthetic signals consistently elevate class-wise performance across all subjects and model families. 
This integration demonstrates the dual value of forecasting-based synthesis: enabling open biomedical datasets and strengthening downstream model robustness. 
Overall, blending synthetic and original time-series data emerges as a practical strategy for building more reliable and generalizable AI in biomedical applications.

\subsection{Visualization of Original vs. Synthetic Data} 
\begin{figure}[ht]
    \centering
    \includegraphics[width=\textwidth]{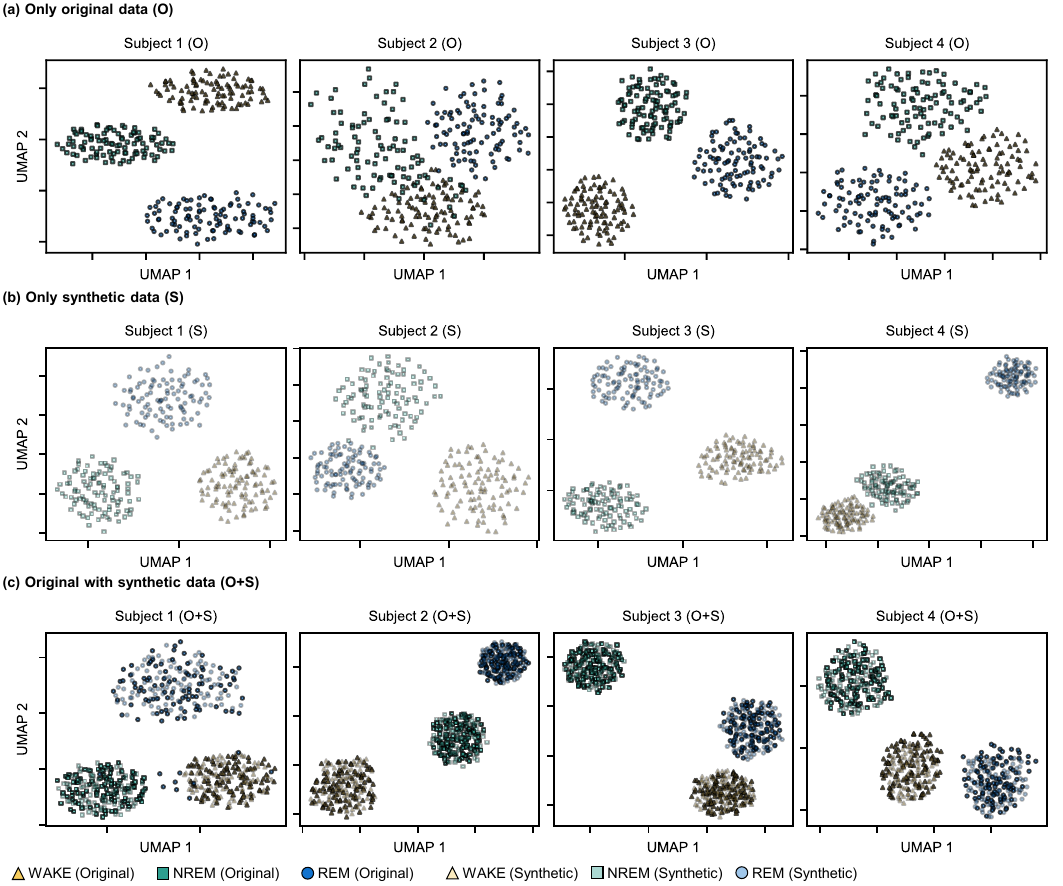}
    \caption{\textbf{UMAP visualization of original and synthetic EEG data across subjects.} The three panels display (a) Original data (O), (b) synthetic only (S), and (c) combined original with synthetic data (O+S) for each subject. Marker shapes denote sleep stages: WAKE (triangles), NREM (squares), and REM (circles).
    }
    \label{fig:5}
\end{figure}
We perform UMAP-based visualization \cite{McInnes2018} to compare the distribution of original and synthetic data across all subjects (Fig. \ref{fig:5}).
The UMAP preserves manifold topology, allowing spatial alignment and inter-class separation to reliably indicate feature fidelity and decision boundary retention, respectively \cite{ghojogh2023elements}.
Note that the synthetic data were generated using each subject’s best-performing forecaster (S / O+S): SOFTS / DeepAR for Subject 1, TiDE / TiDE for Subject 2, KAN / TiDE for Subject 3, and SOFTS / DLinear for Subject 4, which are consistently applied in all subsequent experiments.
For each subject, 100 samples per class were projected into a 2D space, revealing distinct clusters for WAKE, NREM, and REM stages. 
In the original data (O), clusters exhibit clear separation across all subjects (Fig. \ref{fig:5}a), indicating well-preserved inter-class distinctions except Subject 2.
Note that slight overlaps between NREM and WAKE were observed in Subject 2, reflecting the inherent low separability between these classes specific to this subject.
Moreover, local neighborhood structures remained consistent, reflecting the intrinsic temporal characteristics of EEG signals. 
These original UMAP projections provide a robust reference for evaluating synthetic data fidelity and preserving class boundaries in biomedical time-series synthesis.

We then visualized the synthetic-only (S) data using the best-performing generator for each subject (Fig. \ref{fig:5}b). 
The Transformer-based models such as SOFTS and TiDE, and forecasters such as KAN, produced synthetic clusters that closely aligned with their original counterparts across all subjects.
In Subjects 1–3, the WAKE and NREM clusters remained compact with minimal drift, indicating that key temporal dynamics were effectively captured.
Notably, Subject 4 demonstrated even tighter clustering across all classes, suggesting that the generator concentrated more heavily on dominant and reliable feature regions.
No spurious clusters or artifacts were observed, confirming the physiological plausibility of the synthetic data.
The results indicate that time-series forecasters can produce synthetic EEG data that respects both global structure and local variations of the original signals.

Finally, we examine UMAP plots with original and synthetic data merged (O+S) (Fig. \ref{fig:5}c). 
In all subjects, synthetic data effectively complemented the original manifold, filling sparse regions and enhancing data coverage. 
Notably, in Subject 1, synthetic REM points extended into low-density areas, addressing feature space gaps. 
Moreover, except for Subject 1, the combined clusters exhibit tighter boundaries, indicating improved class balance and coverage.
Thus, the insights underscore the utility of synthetic data in enhancing diversity and generalization for downstream biomedical AI tasks.
\subsection{Evaluation of Quality and Privacy Metrics}
\begin{table}[t]
\centering
\scriptsize
\setlength{\tabcolsep}{4pt}% tighter columns
\caption{\textbf{Quantitative assessment of synthetic data quality and privacy preservation across subjects.} The table presents quality metrics and privacy metrics for Synthetic-only (S) and Combined (O+S) modes.}
\label{tab:2}
\begin{tabular}{llcccc}
\toprule
\multirow{2}{*}{\textbf{Subject}} & \multirow{2}{*}{\textbf{Mode}} & \multicolumn{2}{c}{\textbf{Quality Metrics}} & \multicolumn{2}{c}{\textbf{Privacy Metrics}} \\
\cmidrule(lr){3-4} \cmidrule(lr){5-6}
 &  & \textbf{Stat Sim.} & \textbf{Corr Sim.} & \textbf{DCR} & \textbf{NNDR} \\
\midrule
\multirow{2}{*}{Subject 1} & S & 0.9667 & 0.9639 & 0.6592 & 0.9569 \\
 & O+S & 0.9731 & 0.9567 & 0.4558 & 0.9759 \\
\midrule
\multirow{2}{*}{Subject 2} & S & 0.9744 & 0.9504 & 0.3619 & 0.9525 \\
 & O+S & 0.9735 & 0.9460 & 0.3570 & 0.9598 \\
\midrule
\multirow{2}{*}{Subject 3} & S & 0.9742 & 0.9747 & 0.4571 & 0.9676 \\
 & O+S & 0.9750 & 0.9866 & 0.4430 & 0.9766 \\
\midrule
\multirow{2}{*}{Subject 4} & S & 0.9711 & 0.9763 & 0.4951 & 0.9266 \\
 & O+S & 0.9779 & 0.9797 & 0.1804 & 0.8942 \\
\bottomrule
\end{tabular}
\end{table}

To evaluate both the quality and privacy of the generated datasets, we conducted a quantitative assessment using the best-performing forecaster for each subject under the synthetic-only (S) and combined (O+S) settings  (Table \ref{tab:2}).
We randomly sampled 50 synthetic instances per condition to compute the quality metrics—Statistical Similarity (Stat Sim.) and Correlation Similarity (Corr Sim.) \cite{yang2024structured}.
Using the same samples, we evaluated the privacy metrics, Distance to Closest Record (DCR) and Nearest Neighbor Distance Ratio (NNDR) \cite{steier2025synthetic}.
The quality metrics show that the synthetic data consistently preserved key temporal and statistical characteristics, with Stat Sim. values exceeding 0.96 across all subjects.
% Moreover, for Subject 4, the S setting achieved the highest Corr Sim. (0.9763), demonstrating that the method effectively captures complex temporal and inter-variable dynamics in EEG signals.

In terms of privacy preservation, we analyzed distance-based metrics to assess the risk of the synthetic samples simply memorizing the training data.
The DCR values were non-zero across all subjects in both the S and O+S settings, indicating measurable separation from the real data and thereby reducing privacy risk.
In addition, the NNDR values remained consistently high across subjects, with Subject 3 reaching 0.9676 in the S mode, indicating strong resistance to nearest-neighbor–based re-identification.
Overall, the results show that the method yields representative data without duplicating the originals.
Note that this evaluation uses a vanilla setting, and adding differential privacy \cite{abadi2016deep} can further strengthen privacy.

\subsection{Comparison with GAN-based Synthesis}

\begin{figure}[t]
    \centering
    \includegraphics[width=0.5\textwidth]{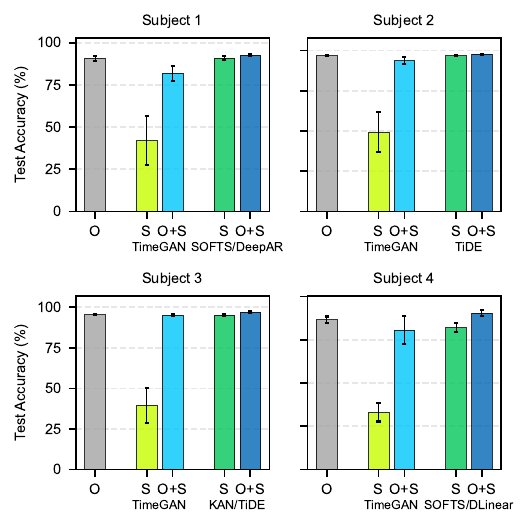}
    \caption{\textbf{Comparison of test accuracy (\%) between synthetic data generated by TimeGAN and by the best-performing time-series forecasters per subject.} The Bars show original-only (O), synthetic-only (S), and combined original with synthetic data (O+S) conditions across subjects. Each value denotes mean ± std computed from five independent seeds.
    }
    \label{fig:7}
\end{figure}
We compare the performance of synthetic data generated by TimeGAN \cite{yoon2019time}, a representative generative model for time-series data, with that of our proposed time-series forecaster-based approach (Fig. \ref{fig:7}). 
Under the synthetic-only (S) condition, our forecaster-based method consistently outperformed TimeGAN across all subjects. 
For instance, in Subject 1, TimeGAN achieved 42.00\%, whereas our method reached 91.00\% with SOFTS. 
Similarly, in Subject 2, TimeGAN yielded 49.47\%, while our TiDE-based approach achieved 96.76\%. 
In Subjects 3 and 4, the gap remained substantial, with our models exceeding 95\% and 82\%, respectively, compared to TimeGAN’s 39.45\% and 32.96\%. These results indicate that our forecasting models more effectively capture temporal dynamics critical for biomedical time-series data. 
Moreover, the low variance observed in our method highlights its robustness across different seeds.

When synthetic data was combined with original data (O+S), our approach continued to deliver superior results. 
In Subject 2, the O+S performance reached 97.71\% with TiDE, surpassing TimeGAN’s 93.77\%. 
Likewise, in Subject 1, the combination of DeepAR and synthetic data achieved 92.75\%, well above TimeGAN’s 81.83\%. 
In Subject 3, both approaches approached original performance, but our method still led with 96.96\%. 
The Subject 4 showed the largest relative gain, with our method reaching 90.31\% compared to TimeGAN’s 80.57\%. 
Notably, across all subjects, our forecaster-based synthesis produced more stable and higher accuracy with lower standard deviations. 
Therefore, the trends demonstrate that our approach generates synthetic data that more effectively complements original datasets, enhancing model performance in biomedical time-series classification.

\section{Discussion}\label{discussion}
\begin{figure}[t]
    \centering
    \includegraphics[width=0.5\textwidth]{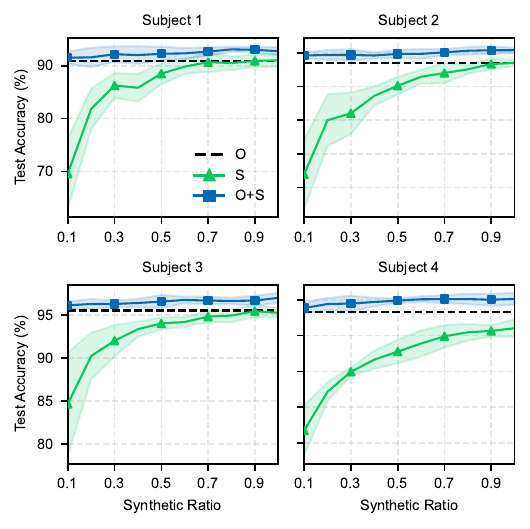}
    \caption{\textbf{Impact of synthetic data ratio on classification accuracy.} Test accuracy for Synthetic-only (S, green triangles) and Combined (O+S, blue squares) conditions is plotted against synthetic proportions (0.1--1.0), relative to the Original-only baseline (O, black dashed line). Shaded regions represent the standard deviation across five independent seeds.}
    \label{fig:6}
\end{figure}

To evaluate the impact of synthetic data volume, we assessed model performance across synthetic-to-original ratios ranging from 0.1 to 1.0 (Fig. \ref{fig:6}). 
In the synthetic-only (S) condition, accuracy was highly sensitive to data volume.
For example, Subject 4 improved from 53.46\% to 82.08\% as the ratio increased, indicating that sufficiently large synthetic datasets are required to capture the underlying distribution.
By contrast, the combined (O+S) condition showed stable robustness and consistent gains over the baseline, yet accuracy tended to saturate beyond a ratio of 0.8. 
This suggests that adopting adaptive generation strategies that adjust synthetic volume based on model feedback may yield additional improvements. 
Note that although this work demonstrates the utility of the proposed framework for EEG signals, further validation is needed to extend it to more complex modalities such as multi-channel or irregular biomedical time-series.
Looking ahead, future work should explore these directions to establish application-specific synthesis protocols that maximize utility while managing computational and privacy costs.

\section{Conclusion}\label{conclusion}
In this work, we demonstrate that forecasting-based synthetic data generation offers a powerful and scalable approach for biomedical time-series applications. 
Our method consistently outperformed conventional GAN-based synthesis, producing high-fidelity data that enhances AI model robustness. 
Moreover, the open-source release of these synthetic datasets will help democratize access to biomedical data while preserving patient privacy. 
Importantly, our results show that combining synthetic with original data improves generalization and enables more reliable clinical AI systems. 
In future work, we aim to develop this framework for specific clinical pathologies and integrate synthetic pipelines into real-world healthcare workflows.

\section*{Acknowledgments}
This research was partly supported by the Institute of Information \& Communications Technology Planning \& Evaluation (IITP)-ITRC (Information Technology Research Center) grant funded by the Korea government (MSIT) (IITP-2026-RS-2020-II201787, contribution rate: 50\%).
Also, this research was supported in part by a grant of the Korea-US Collaborative Research Fund (KUCRF), funded by the Ministry of Science and ICT and Ministry of Health \& Welfare, Republic of Korea (grant number: RS-2025-16022980, contribution rate: 50\%).

\section*{Ethics Statement}
This work did not involve any new animal or human experiments.

\bibliographystyle{elsarticle-num}
\bibliography{reference}

\end{document}